\documentclass{article} 
\usepackage{iclr2019_spirl_workshop,times}

\usepackage{amsmath,amsfonts,bm}

\def\eqref#1{equation~\ref{#1}}

\def\1{\bm{1}}

\def\ra{{\textnormal{a}}}

\def\rs{{\textnormal{s}}}

\def\va{{\bm{a}}}

\def\vo{{\bm{o}}}

\def\vs{{\bm{s}}}

\def\vz{{\bm{z}}}

\DeclareMathAlphabet{\mathsfit}{\encodingdefault}{\sfdefault}{m}{sl}
\SetMathAlphabet{\mathsfit}{bold}{\encodingdefault}{\sfdefault}{bx}{n}

\newcommand{\E}{\mathbb{E}}

\newcommand{\KL}{D_{\mathrm{KL}}}

\usepackage{hyperref}
\usepackage{url}
\usepackage{subcaption}
\usepackage{appendix}
\usepackage{todonotes}
\usepackage{amssymb}
\usepackage{amsmath}

\newcommand{\bt}[1]{\widetilde{\boldsymbol{#1}}}

\title{Bayesian policy selection \\ using Active Inference}

\author{Ozan \c{C}atal, Johannes Nauta, Tim Verbelen, Pieter Simoens, \& Bart Dhoedt\\
IDLab, Department of Information Technology \\
Ghent University - imec\\
Ghent, Belgium \\
\texttt{ozan.catal@ugent.be}
}

\iclrfinalcopy 
\begin{document}

\maketitle

\begin{abstract}
   Learning to take actions based on observations is a core requirement for artificial agents to be able to be successful and robust at their task. Reinforcement Learning (RL) is a well-known technique for learning such policies. However, current RL algorithms often have to deal with reward shaping, have difficulties generalizing to other environments and are most often sample inefficient.
   In this paper, we explore active inference and the free energy principle, a normative theory from neuroscience that explains how self-organizing biological systems operate by maintaining a model of the world and casting action selection as an inference problem. We apply this concept to a typical problem known to the RL community, the mountain car problem, and show how active inference encompasses both RL and learning from demonstrations.
\end{abstract}

\section{Introduction}
\label{intro}
Active inference is an emerging paradigm from neuroscience~\citep{Friston2017}, which postulates that action selection in biological systems, in particular the human brain, is in effect an inference problem where agents are attracted to a preferred prior state distribution in a hidden state space. Contrary to many state-of-the art RL algorithms, active inference agents are not purely goal directed and exhibit an inherent epistemic exploration \citep{Schwartenbeck2018}. In neuroscience, the idea of using active inference~\citep{Friston2010,Friston2006} to solve different control and learning tasks has already been explored~\citep{Friston2012a, Friston2013, Friston2017}. These approaches however make use of manually engineered transition models and predefined, often discrete, state spaces.

In this work we make a first step towards extrapolating this inference approach to action selection by artificial agents by the use of neural networks for learning a state space, as well as models for posterior and prior beliefs over these states. We demonstrate that by minimizing the variational free energy a dynamics model can be learned from the problem environment, which is sufficient to reconstruct and predict environment observations. This dynamics model can be leveraged to perform inference on possible actions as well as to learn habitual policies.

\section{Active Inference \& the free energy principle}
\label{ai}
Free energy is a commonly used quantity in many scientific and engineering disciplines, describing
the amount of work a (thermodynamic) system can perform. Physical systems will always move towards a state of minimal free energy. In active inference the concept of variational free energy is utilized to describe the drive of organisms to self-organisation. The free energy principle states that every organism entertains an internal model of the world, and implicitly tries to minimize the difference between what it believes about the world and what it perceives, thus minimizing its own variational free energy~\citep{Friston2010}, or alternatively the Bayesian surprise.
Concretely this means that every organism or agent will actively drive itself towards preferred world states, a kind of global prior, that it believes a priori it will visit. In the context of learning to act, this surprise minimization boils down to two distinct objectives. On the one hand the agent actively samples the world to fine tune its internal model of the world and better explain observations. On the other hand the agent will be driven to visit preferred states which carry little expected free energy.

Formally, an agent entertains a generative model $P(\tilde{\vo},\tilde{\va},\tilde{\vs})$ of the environment, which specifies the joint probability of observations, actions and their hidden causes, where actions are determined by some policy $\pi$. The reader is referred to Appendix \ref{Appendix:Glossary} for an overview of the used notation.

If the environment is modelled as a Markov Decision Process (MDP) this generative model factorizes as:
\begin{equation}
   \label{eq:factor}
   P(\tilde{\vo}, \tilde{\va}, \tilde{\vs}) = P(\pi)P(\vs_0)\prod_{t=1}^{T}P(\vo_t|\vs_t)P(\vs_{t}|\vs_{t-1}, \va_{t})P(\va_{t}|\pi)
\end{equation}

The free energy or Bayesian surprise is then defined as:
\begin{equation}
   \label{eq:free-energy}
   \begin{split}
      F &= \E_Q [ \log Q(\tilde{\vs}) - \log P(\tilde{\vs},\tilde{\vo} )] \\
        &= \KL (Q(\tilde{\vs}) \Vert P(\tilde{\vs} | \tilde{\vo}) ) - \log P(\tilde{\vo}) \\
        &= \KL (Q(\tilde{\vs}) \Vert P(\tilde{\vs})) - \E_{Q} [ \log P(\tilde{\vo} | \tilde{\vs})]
   \end{split}
\end{equation}

where $Q(\tilde{\vs})$ is an approximate posterior distribution. The second equality shows that the free energy is minimized when the KL divergence term becomes zero, meaning that the approximate posterior becomes the true posterior, in which case the free energy becomes the negative log evidence. The third equality then becomes the negative evidence lower bound (ELBO), as we know from variational autoencoders (VAE) \citep{Kingma13, Rezende14}.
For a complete derivation the reader is referred to Appendix \ref{Appendix:Free-energy}.

In active inference agents pick actions that will result in visiting states of low expected free energy. Concretely, agents do this by sampling actions from a prior belief about policies according to how much expected free energy that policy will induce. According to \cite{Schwartenbeck2018} this means that the probability of picking a policy is given by
\begin{equation}
   \label{eq:G}
   \begin{split}
      P(\pi) &= \sigma(-\gamma G (\pi)) \\
      G(\pi) &= \sum_{\tau}^{T} G(\pi, \tau)
   \end{split}
\end{equation}
where $\sigma$ is the softmax function with precision parameter $\gamma$, which governs the agent’s goal-directedness and randomness in its behavior. $G$ is the expected free energy at future time-step $\tau$  under policy $\pi$, which can be expanded into:
\begin{equation}
   \label{eq:G-complete}
   \begin{split}
      G(\pi, \tau) &= \E_{Q(\vo_{\tau}, \vs_{\tau} \vert \pi)} [ \log Q(\vs_{\tau} \vert \pi) - \log P(\vo_{\tau}, \vs_{\tau} \vert \pi)] \\
                   &= \E_{Q(\vo_{\tau}, \vs_{\tau} \vert \pi)} [ \log Q(\vs_{\tau} \vert \pi) - \log P(\vo_\tau \vert \vs_\tau, \pi) - \log P(\vs_\tau \vert \pi)] \\
                   &= \KL (Q(\vs_\tau | \pi) \Vert P(\vs_\tau)) + \E_{Q(\vs_\tau)} [ H(P(\vo_\tau \vert \vs_\tau))]
  \end{split}
\end{equation}
We used $Q(\vo_\tau,\vs_\tau \vert \pi) = P(\vo_\tau \vert \vs_\tau)Q(\vs_\tau \vert \pi)$ and that the prior probability $P(\vs_\tau \vert \pi)$ is given by a preferred state distribution $P(\vs_\tau)$. This results into two terms: a KL divergence term between the predicted states and the prior preferred states, and an entropy term reflecting the expected ambiguity under predicted states. Action selection in active inference thus entails:
\begin{enumerate}
   \item Evaluate $G(\pi)$ for each policy $\pi$
   \item Calculate the belief over policies $P(\pi)$
   \item Infer the next action using $P(\pi)P(\va_{t+1} \vert \pi)$
\end{enumerate}
However, examples applying this principle are often limited to cases with a discrete number of predefined policies, as otherwise calculating $P(\pi)$ becomes intractable \citep{Friston2017}.

\section{Neural networks as density estimators}
\label{nn}
In order to overcome the intractability of calculating $P(\pi)$ we characterize the approximate posterior with a neural network with parameters $\phi$ according to the following factorization:
\begin{equation*}
   Q(\tilde{\vs}) = \prod_{t=1}^{T} q_{\phi}(\vs_{t} | \vs_{t-1}, \va_{t}, \vo_{t})
\end{equation*}
Similarly we parameterise a likelihood model $p_{\xi}(\vo_{t}|\vs_{t})$ and dynamics model $p_{\theta}(\rs_t | \rs_{t-1}, \ra_{t-1})$ as neural networks with parameters $\xi$ and $\theta$. These networks output a multivariate Gaussian distribution with diagonal covariance matrix using the reparametrization trick from \cite{Kingma13}.

Minimizing the free energy then boils down to minimizing the objective:
\begin{equation}
   \label{eq:model-obj}
   \forall t : \underset{\phi, \theta, \xi}{\text{minimize}}: -\log p_{\xi}(\vo_{t}|\vs_{t})
   + \KL (q_{\phi}(\vs_t | \vs_{t-1}, \va_{t}, \vo_t) \Vert p_{\theta}(\vs_t | \vs_{t-1}, \va_{t}))
\end{equation}

The negative log likelihood term of the objective punishes reconstruction error, forcing all information from the observations into the state space. The KL term pulls the prior distribution, or the state transition model, to the posterior model, also known as the observation model, forcing it to learn to encode state distributions from which observations can be reconstructed without having actual access to these observations. This can be interpreted as a variational autoencoder (VAE), where instead of a global prior, the prior is given by the state transition model.

Action selection is then realized by using the state transition model to sample future states, given a sequence of (randomly sampled) actions. For each action sequence we evaluate $G$ and we execute the first action of the sequence with minimal $G$. Any random sampling strategy can be used, for example the cross entropy method~\cite{Rubinstein96}.

The cumbersome sampling according to expected free energy can be avoided by using amortized inference. In this case we also instantiate a ``habit'' policy $P(\va_t \vert \vs_t)$ as a neural network that maps states to actions in a deterministic or stochastic way. We can train this neural network using back propagation by minimizing $G$.

\begin{figure}[t!]
   \centering
   \includegraphics[width=5in]{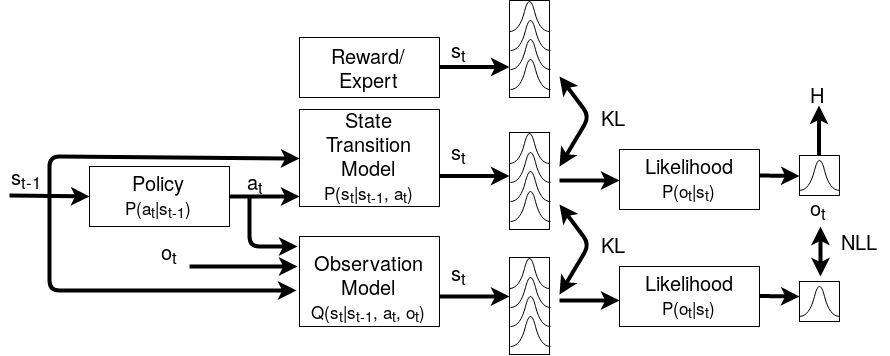}
   \caption{The various components of the agent and their corresponding training losses.
   We minimize the variational free energy by minimizing both the negative log likelihood of observations and the KL divergence between the state transition model and the observation model. The inferred hidden state is characterized as a multivariate Gaussian distribution. Policy learning is achieved by minimizing the expected free energy $G$ between the state distribution visited by the policy according to the State Transition Model and the state distribution visited by the expert.}
   \label{fig:training}
\end{figure}

Crucially, we still need to define the preferred states or global prior $P(\vs_\tau)$. When we have access to a reward signal, this can be converted to a preferred state prior by putting more probability density on rewarding states. Otherwise, we can initialize the system with a flat prior (each state is equally preferred), and increase the probability of rewarding states as we visit them. When we have access to an expert, we can use expert demonstrations to provide a prior on preferred states, i.e. the states visited by the expert policy. An overview of the various models and train losses is shown in Figure~\ref{fig:training}.

\section{Experiments}
\label{results}
We validate our theoretical framework on the continuous mountain car problem from the OpenAI Gym~\citep{gym}, adapted to only provide noisy observations (and no access to the velocity state). This environment, although quite simple in terms of problem complexity, proves a good initial trial problem due to greedy approaches failing at it.

\begin{figure}[t!]
    \centering
    \begin{subfigure}[t]{0.49\textwidth}
        \centering
        \includegraphics[width=0.9\textwidth]{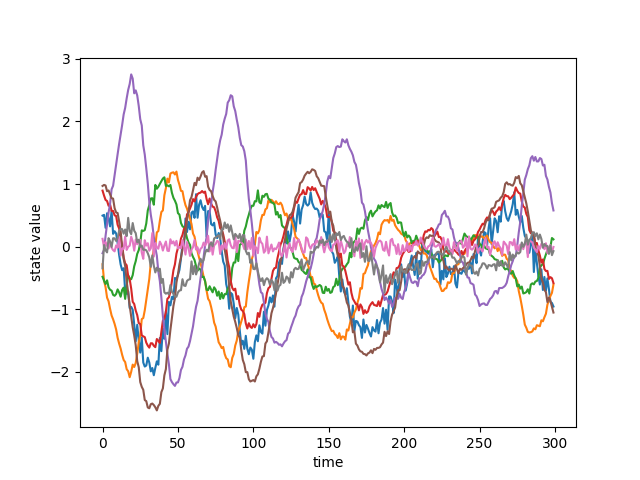}
        \caption{Resulting state space during a role-out.}
        \label{fig:states}
    \end{subfigure}
    \begin{subfigure}[t]{0.49\textwidth}
        \centering
        \includegraphics[width=0.9\textwidth]{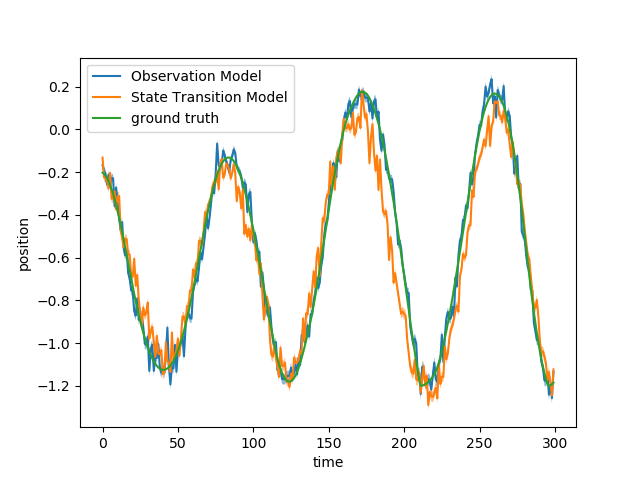}
        \caption{Predicted observations plotted on top of the ground truth observations.}
        \label{fig:predictions}
    \end{subfigure}
    \caption{Results of the model training stage. In Figure (a) we see that every parameter of the eight dimensional state space encodes some element contributing to the prediction.
    Figure (b) shows that the models are capable of predicting ground truth observations, indicating that they accurately learned the environment dynamics.
    }
    \label{fig:model-training}
\end{figure}

We instantiate the state transition model, observation model and likelihood models as fully connected neural networks with 64 hidden units. The agent's internal state is parameterised as an 8 dimensional multivariate Gaussian $\vs$. We bootstrap the model by training on a random agent and optimizing Eq.~\ref{eq:model-obj}. Our random agent samples actions uniformly in $[-1,1]$, with a 90\% chance of repeating the previous action. Figure~\ref{fig:states} shows a plot of the evolution of the means of the internal states during a random rollout, whilst Figure~\ref{fig:predictions} shows reconstructions from both the state transition model and observation model from the same rollout. Note that the state transition model has no access to any observations. The closeness between ground truth observations and state transition model reconstructions illustrates that the agent has successfully learned an internal world representation sufficient to predict the world evolution from a single initial observation.

\begin{figure}[t!]
    \centering
    \begin{subfigure}[t]{0.45\textwidth}
        \includegraphics[width=0.9\textwidth]{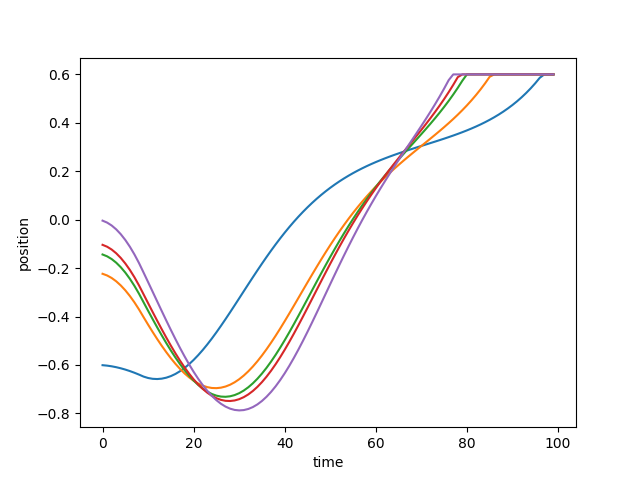}
        \caption{Expert roll-outs from 5 different start positions.}
        \label{fig:expert}
    \end{subfigure}\quad
    \begin{subfigure}[t]{0.45\textwidth}
        \includegraphics[width=0.9\textwidth]{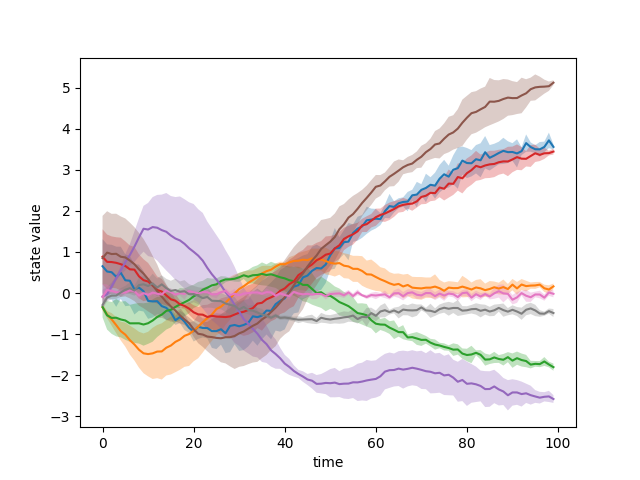}
        \caption{Preferred state distribution from expert roll-outs. Each curve indicates the distribution of a latent dimension during rollout.}
        \label{fig:expert-states}
    \end{subfigure}
    \begin{subfigure}[t]{0.45\textwidth}
        \includegraphics[width=0.9\textwidth]{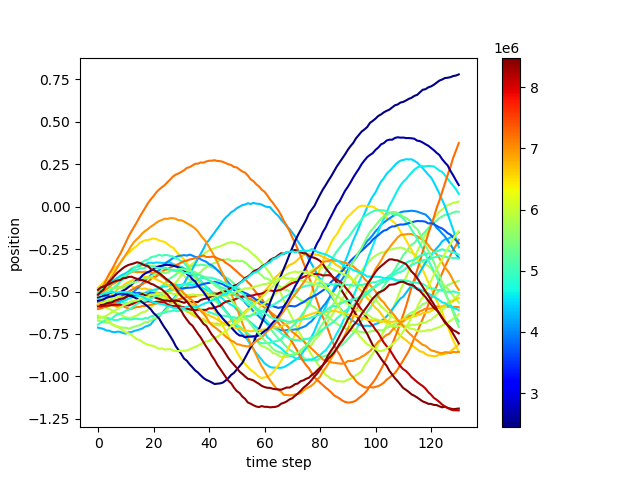}
        \caption{Random imaginary rollouts with expected free~energy G, represented in the color bar. Lower G is better.}
        \label{fig:g-rollouts}
    \end{subfigure}\quad
    \begin{subfigure}[t]{0.45\textwidth}
        \includegraphics[width=0.9\textwidth]{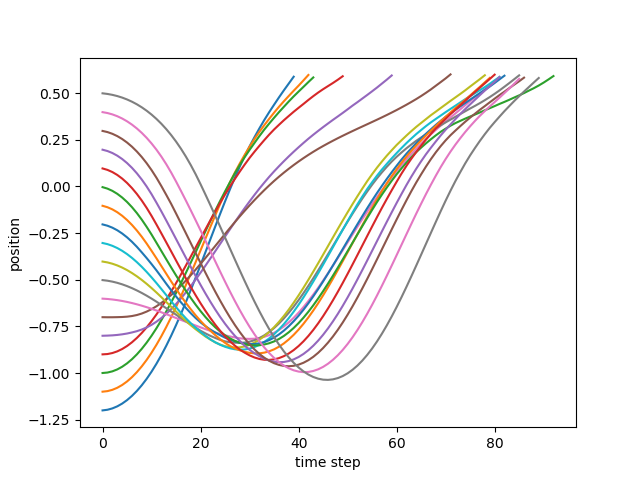}
        \caption{Actual policy rollouts with different starting positions.}
        \label{fig:apprentice-policy}
    \end{subfigure}
    \caption{Results of the policy learning. From expert rollouts (a) we distill a preferred state distribution (b). Sampling imaginary rollouts of the state transition model results in better rollouts having lower expected free energy G (c). We can use this sampling to train an amortized active inference policy that successfully solves the environment from any starting position (d).}
    \label{fig:policy-learning}
\end{figure}

To construct a preferred state prior, we manually execute 5 ``expert rollouts'' in the environment. These rollouts can be seen in Figure~\ref{fig:expert}. From these rollouts a preferred state distribution is extracted (Figure~\ref{fig:expert-states}). You can see that in the beginning of the sequence, there is a variance on the preferred states, whereas towards the end the preferred state distribution is peaked around the state reflecting the car being on the top of the mountain. This is equivalent with a sparse reward signal when reaching the mountain top at the end of the sequence. Similarly, one can also engineer a preferred state prior based on the reward signal, which we discuss in Appendix~\ref{Appendix:Rewards}

We now use this state policy for action selection according the active inference scheme. We sample random rollouts using the state transition model and calculate the expected free energy G for each. This indeed selects rollouts that successfully reach the mountain top as shown in Figure~\ref{fig:g-rollouts}. Next, we also train a policy by minimizing $G$ at every timestep as defined in Eq~\ref{eq:G-complete}. After training, this policy is indeed able to generalize to any starting position, consistently reaching the mountain top.

\section{Related Work}
\label{related}
Model-free RL has been successfully proven to work on many game playing~\citep{Silver2016,Hessel2017} and robotics problems~\citep{Yahya2017, Kalashnikov2018}. However, there are still some outstanding challenges, such as the issue of reward engineering~\citep{Popov17}, generalizing to other environments~\citep{Lanctot17}, and sample inefficiency~\citep{Yu2018}.

Recently there have been promising advancements in the area of model-based RL. Ha et al. \cite{Ha2018} train a variational autoencoder (VAE) in conjunction with a recurrent dynamics model to create a predictive world model. They use this predictive model to train a controller using evolution strategies~\citep{Salimans2017}. \cite{Bohez2018} train jointly a prior and posterior model for a robotics navigation task. In MERLIN~\citep{Wayne2018}, a neural memory module is added on top of a VAE based dynamics model to facilitate learning policies over longer time windows. In a similar vein \citep{Srinivas2018} learns abstract representation for planning.

Instead of using an explicit reward signal, policies can also be learned from demonstrations using inverse reinforcement learning, basically learning a reward signal from an expert~\citep{Ng2000}. Another approach for learning from demonstrations is using meta-learning to quickly distill a policy from a new demonstration \citep{Finn2017}.

Active inference combines elements of all these works into a single theoretical framework. The preferred states prior can be interpreted as more generic form of value function in RL. It combines learning world models with action selection and planning. Also, the concept of minimizing expected Bayesian surprise resembles the work on artificial curiosity for exploration~\citep{Schmidhuber2011}.
\section{Conclusion}
\label{discussion}
Active inference might underlie the way biological agents perceive and interact with the world. In this work we show that active inference and the free energy principle can also help artificial agents interact with the world. Active inference also combines a lot of elements from recent RL literature, such as building world models, neural planning, artificial curiosity, etc. We belief this is an interesting direction for further research, to apply this principle to more challenging environments, such as the ATARI domain or robotics.

\subsubsection*{Acknowledgments}
Ozan Catal is funded by a Ph.D. grant of the Flanders Research  Foundation (FWO).

\bibliography{main}
\bibliographystyle{iclr2019_spirl_workshop}

\newpage
\appendix
\appendixpage
\section{Glossary}
\label{Appendix:Glossary}
\begin{table}[h!]
   \centering
   \begin{tabular}{c|p{7cm}}
   Definition  &   Description   \\
   \hline
   $\vs_t$ &   State at time $t$ \\
   $\vo_t$ &   Observation at time $t$ \\
   $\va_t$ &   Action at time $t$ \\
   $\vs_{\tau}$ &   Expected state at a future timestep $\tau$ \\
   $\tilde{\vs}$ & Sequence of states \\
   $\tilde{\vo}$ & Sequence of observations \\
   $\tilde{\va}$ & Sequence of actions \\
   $P(\tilde{\vo},\tilde{\va},\tilde{\vs})$ & Generative model of the agent \\
   $P(\vo_t \vert \vs_t)$ & Likelihood model \\
   $\pi$ & Policy \\
   $P(\vs_t \vert \vs_{t-1},\va_{t-1})$ & State transition model \\
   $P(\va_{t} \vert \pi)$ & Action at time $t$ given a policy \\
   $P(\pi)$ & Belief over policies \\
   $P(\tilde{\vs} \vert \tilde{\vo})$  & True posterior about hidden states given a sequence of observations \\
   $Q(\tilde{\vs})$ & Approximate posterior about hidden states \\
   $F = \E_{Q(s)} [\log Q(\tilde{\vs}) - \log P(\tilde{\vs},\tilde{\vo})]$ & Free energy \\
   $G(\cdot)$  & Expected free energy \\
   $\sigma(\vz)_j = e^{z_j} / \sum_k e^{z_k}$  & Softmax or Boltzmann distribution \\
   $\gamma$  & Precision parameter governing goal-directedness and randomness
   \end{tabular}
   \caption{Glossary table with brief description of the used terms and their definitions}
 \end{table}
\newpage
\section{Free energy derivation}
\label{Appendix:Free-energy}
Starting from the definition of free energy:
\begin{equation*}
  F = \E_{Q} [\log Q(\bt{s}) - \log P(\bt{s},\bt{o})]
\end{equation*}

where $Q(\bt{x})$ is an approximate posterior distribution. We can expand the free energy $F$ as follows:
\begin{align*}
  F &= \E_{Q} [ \log Q(\bt{s}) - \log P(\bt{s},\bt{o}) ] \\
  &= \E_{Q} [ \log Q(\bt{s}) - \log P(\bt{s}|\bt{o}) - \log P(\bt{o}) ] \\
  &= \KL (Q(\bt{s}) \Vert P(\bt{x}\vert \bt{o})) - \log P(\bt{o}),
\end{align*}
Similarly, we can also rewrite the free energy expression as:
\begin{align*}
  F &= \E_{Q} [ \log Q(\bt{s}) - \log P(\bt{s},\bt{o}) ] \\
  & \quad  \text{using the identity \;} P(\bt{s},\bt{o}) = P(\bt{o}|\bt{s})P(\bt{s}) \\
  &= \E_{Q}[\log Q(\bt{s}) - \log P(\bt{s}) - \log P(\bt{o} \vert \bt{s})] \\
  &= \KL(Q(\bt{s}) \Vert P(\bt{s})) - \E_{Q} [ \log P(\bt{o}|\bt{s})]
\end{align*}

\section{A reward-based preferred state prior}
\label{Appendix:Rewards}
\begin{figure}[h!]
    \centering
    \includegraphics[width=.6\textwidth]{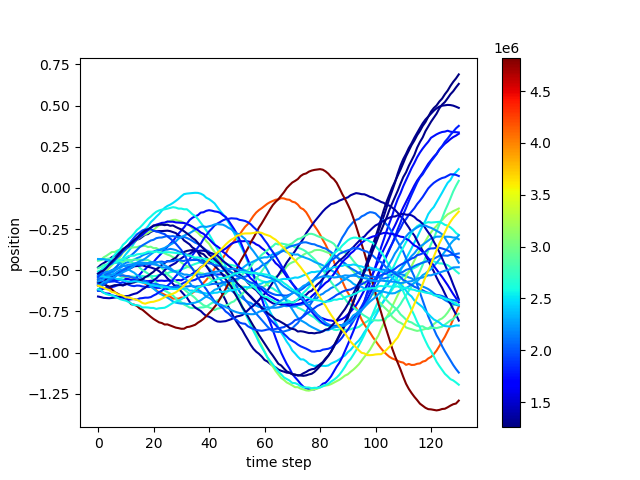}
    \caption{Random imaginary rollouts with corresponding expected free energy G when the preferred state is defined as those states that give reward. Lower G is better.}
    \label{fig:reward}
\end{figure}

In a traditional RL setting, an agent only has access to a reward signal, rather than expert demonstrations. In this case the agent will need to find the preferred state prior that matches the reward signal (higher probability for states that yield higher reward), which is equivalent to learning a value function $V(\cdot)$ in RL. In the mountain car problem, only a sparse reward of +1 is given when reaching the top of the mountain. Based on this reward function, we can define the preferred state distribution as a Gaussian centered around the states where the car reaches the top of the mountain after timestep 100. We can use this prior instead of the one induced by the demonstrations for active inference. We see in Figure \ref{fig:reward} that again trajectories reaching the top result in the lowest expected free energy.
\end{document}